\documentclass[11pt,a4paper]{article}
\usepackage[hyperref]{acl2018}
\usepackage{times}
\usepackage{latexsym}
\usepackage{graphicx}
\usepackage{tabularx}
\usepackage{balance}

\usepackage{amsmath}
\usepackage{breqn}

\usepackage{url}

\aclfinalcopy 
\def\aclpaperid{2} 

\title{A Shared Attention Mechanism for Interpretation\\of Neural Automatic Post-Editing Systems}

\author{Inigo Jauregi Unanue\textsuperscript{1,2}, Ehsan Zare Borzeshi\textsuperscript{2}, Massimo Piccardi\textsuperscript{1} \\
\textsuperscript{1} University of Technology Sydney, Sydney, Australia \\
\textsuperscript{2} Capital Markets Cooperative Research Centre, Sydney, Australia \\
\\
{\tt \{ijauregi,ezborzeshi\}@cmcrc.com, \tt massimo.piccardi@uts.edu.au}
}

\date{}

\begin{document}
\maketitle
\begin{abstract}
Automatic post-editing (APE) systems aim to correct the systematic errors made by machine translators. In this paper, we propose a neural APE system that encodes the source (\textit{src}) and machine translated (\textit{mt}) sentences with two separate encoders, but leverages a shared attention mechanism to better understand how the two inputs contribute to the generation of the post-edited (\textit{pe}) sentences. Our empirical observations have showed that when the \textit{mt} is incorrect, the attention shifts weight toward tokens in the \textit{src} sentence to properly edit the incorrect translation. The model has been trained and evaluated on the official data from the WMT16 and WMT17 APE IT domain English-German shared tasks. Additionally, we have used the extra 500K artificial data provided by the shared task. Our system has been able to reproduce the accuracies of systems trained with the same data, while at the same time providing better interpretability.
\end{abstract}

\section{Introduction}\label{introduction}

In current professional practice, translators tend to follow a two-step approach: first, they run a machine translator (MT) to obtain a first-cut translation; then, they manually correct the MT output to produce a result of adequate quality. The latter step is commonly known as \textit{post-editing} (PE). Stemming from this two-step approach and the recent success of deep networks in MT \cite{sutskever2014,bahdanau2014,luong2015}, the MT research community has devoted increasing attention to the task of automatic post-editing (APE) \cite{bojar2017}.

The rationale of an APE system is to be able to automatically correct the systematic errors made by the MT and thus dispense with or reduce the work of the human post-editors. The data for training and evaluating these systems usually consist of triplets (\textit{src}, \textit{mt}, \textit{pe}), where \textit{src} is the sentence in the source language, \textit{mt} is the output of the MT, and \textit{pe} is the human post-edited sentence. Note that the \textit{pe} is obtained by correcting the \textit{mt}, and therefore these two sentences are closely related. An APE system is ``monolingual'' if it only uses the \textit{mt} to predict the post-edits, or ``contextual'' if it uses both the \textit{src} and the \textit{mt} as inputs \cite{bechara2011}.

Despite their remarkable progress in recent years, neural APE systems are still elusive when it comes to interpretability. In deep learning, highly interpretable models can help researchers to overcome outstanding issues such as learning from fewer annotations, learning with human-computer interactions and debugging network representations \cite{zhang2018visual}. More specifically in APE, a system that provides insights on its decisions can help the human post-editor to understand the system's errors and consequently provide better corrections. As our main contribution, in this paper we propose a contextual APE system based on the seq2seq model with attention which allows for inspecting the role of the \textit{src} and the \textit{mt} in the editing. We modify the basic model with two separate encoders for the \textit{src} and the \textit{mt}, but with a single attention mechanism shared by the hidden vectors of both encoders. At each decoding step, the shared attention has to decide whether to place more weight on the tokens from the \textit{src} or the \textit{mt}. In our experiments, we clearly observe that when the \textit{mt} translation contains mistakes (word order, incorrect words), the model learns to shift the attention toward tokens in the source language, aiming to get extra ``context'' or information that will help to correctly edit the translation. Instead, if the \textit{mt} sentence is correct, the model simply learns to pass it on word by word. In Section \ref{results}, we have plotted the attention weight matrices of several predictions to visualize this finding.

The model has been trained and evaluated with the official datasets from the WMT16 and WMT17 Information Technology (IT) domain APE English-German (en-de) shared tasks \cite{bojar2016,bojar2017}. We have also used the 500K artificial data provided in the shared task for extra training. For some of the predictions in the test set, we have analysed the plots of attention weight matrices to shed light on whether the model relies more on the \textit{src} or the \textit{mt} at each time step. Moreover, our model has achieved higher accuracy than previous systems that used the same training setting (official datasets + 500K extra artificial data).

\section{Related work}\label{related_work}

In an early work, \cite{simard2007} combined a rule-based MT (RBMT) with a statistical MT (SMT) for monolingual post-editing. The reported results outperformed both systems in standalone translation mode. In 2011, \cite{bechara2011} proposed the first model based on contextual post-editing, showing improvements over monolingual approaches. 

More recently, neural APE systems have attracted much attention. \cite{junczys2016} (the winner of the WMT16 shared task) integrated various neural machine translation (NMT) components in a log-linear model. Moreover, they suggested creating artificial triplets from out-of-domain data to enlarge the training data, which led to a drastic improvement in PE accuracy. Assuming that post-editing is reversible, \cite{pal2017} have proposed an attention mechanism over bidirectional models, \textit{mt}$\rightarrow$ \textit{pe} and \textit{pe} $\rightarrow$ \textit{mt}. Several other researchers have proposed using multi-input seq2seq models for contextual APE \cite{berard2017,libovicky2016,varis2017, pal2017, libovicky2017,chatterjee2017}. All these systems employ separate encoders for the two inputs, \textit{src} and \textit{mt}.

\subsection{Attention mechanisms for APE}

A key aspect of neural APE systems is the attention mechanism. A conventional attention mechanism for NMT first learns the  alignment scores ($e^{ij}$) with an alignment model \cite{bahdanau2014,luong2015} given the $j$-th hidden vector of the encoder ($\textbf{h}^{j}$) and the decoder's hidden state ($\textbf{s}_{i-1}$) at time $i-1$ (Equation \ref{eq:alignScore}). Then, Equation \ref{eq:softmax} computes the normalized attention weights, with $T_x$ the length of the input sentence. Finally, the context vector is computed as the sum of the encoder's hidden vectors weighed by the attention weights (Equation \ref{eq:context_usually}). The decoder uses the computed context vector to predict the output.

\vspace{-6pt}

\begin{equation}
\label{eq:alignScore}
\begin{split}
 e^{ij}=aligment\_model(\textbf{h}^{j},\textbf{s}^{i-1}) 
\end{split}
\end{equation}

\vspace{-18pt}

\begin{equation}
\label{eq:softmax}
\begin{split}
\alpha^{ij} = \dfrac{exp(e^{ij})}{\sum_{m=1}^{T_x}exp(e^{im})}
\end{split}
\end{equation}

\vspace{-18pt}

\begin{equation}
\label{eq:context_usually}
\begin{split}
\textbf{c}^{i}=\sum_{j=1}^{T_x}\alpha^{i,j}\textbf{h}^j 
\end{split}
\end{equation}

In the APE literature, two recent papers have extended the attention mechanism to contextual APE. \cite{chatterjee2017} (the winner of the WMT17 shared task) have proposed a two-encoder system with a separate attention for each encoder. The two attention networks create a context vector for each input, $\textbf{c}_{src}$ and $\textbf{c}_{mt}$, and concatenate them using additional, learnable parameters, $\textbf{\textit{W}}_{ct}$ and $\textbf{b}_{ct}$, into a merged context vector, $\textbf{c}_{merge}$ (Equation \ref{eq:context_concat}).

\vspace{-6pt}

\begin{equation}
\label{eq:context_concat}
\begin{split}
\textbf{c}_{merge}^{i}=[{\textbf{c}_{src}^{i}};{\textbf{c}_{mt}^{i}}]*\textbf{\textit{W}}_{ct} +\textbf{b}_{ct} 
\end{split}
\end{equation}

\cite{libovicky2017} have proposed, among others, an attention strategy named the \textit{flat attention}. In this approach, all the attention weights corresponding to the tokens in the two inputs are computed with a joint soft-max:

\vspace{-6pt}

\begin{equation}
\label{eq:joint_softmax}
\begin{split}
\alpha_{(k)}^{ij} = \dfrac{exp(e_{(k)}^{ij})}{\sum_{n=1}^{2}\sum_{m=1}^{T_x^{(n)}}exp(e_{(n)}^{im})}
\end{split}
\end{equation}

\noindent where $e^{ij}_{(k)}$ is the attention energy of the $j$-th step of the $k$-th encoder at the $i$-th decoding step and $T_x^{(k)}$ is the length of the input sequence of the $k$-th encoder. Note that because the attention weights are computed jointly over the different encoders, this approach allows observing whether the system assigns more weight to the tokens of the \textit{src} or the \textit{mt} at each decoding step. Once the attention weigths are computed, a single context vector ($\textbf{c}$) is created as:

\vspace{-18pt}

\begin{equation}
\label{eq:flat_attention_context}
\begin{split}
\textbf{c}^{i}=\sum_{k=1}^{N}\sum_{j=1}^{T_x^{(k)}}\alpha_{(k)}^{i,j}\textbf{U}_{c(k)}\textbf{h}_{(k)}^j
\end{split}
\end{equation}

\noindent where $\textbf{h}_{(k)}^j$ is the $j$-th hidden vector from the $k$-th encoder, $T_x^{(k)}$ is the number of hidden vectors from the $k$-th encoder, and $\textbf{U}_{c(k)}$ is the projection matrix for the $k$-th encoder that projects its hidden vectors to a common-dimensional space. This parameter is also learnable and can further re-weigh the two inputs.

\section{The proposed model}\label{the_model}

The main focus of our paper is on the interpretability of the predictions made by neural APE systems. To this aim, we have assembled a contextual neural model that leverages two encoders and a shared attention mechanism, similarly to the \textit{flat attention} of \cite{libovicky2017}. To describe it, let us assume that $\textbf{X}_{src}=\{\textbf{x}_{src}^1, ... , \textbf{x}_{src}^N\}$ is the \textit{src} sentence and $\textbf{X}_{mt}=\{\textbf{x}_{mt}^1, ... , \textbf{x}_{mt}^M\}$ is the \textit{mt} sentence, where $N$ and $M$ are their respective numbers of tokens. The two encoders encode the two inputs separately:

\begin{equation}
\label{eq:encoders}
\begin{split}
\textbf{h}_{src}^{j}=enc_{src}(\textbf{x}_{src}^j,\textbf{h}_{src}^{j-1})  \quad j=1,...,N\\
\textbf{h}_{mt}^{j}=enc_{mt}(\textbf{x}_{mt}^j,\textbf{h}_{mt}^{j-1})  \quad j=1,...,M 
\end{split}
\end{equation}

All the hidden vectors outputs by the two encoders are then concatenated as if they were coming from a single encoder:

\begin{equation}
\label{eq:joinHidden}
\textbf{h}_{join}=\{\textbf{h}_{src}^1, ... , \textbf{h}_{src}^N, \textbf{h}_{mt}^{1}, ..., \textbf{h}_{mt}^{M}\}\\ 
\end{equation}

Then, the attention weights and the context vector at each decoding step are computed from the hidden vectors of $\textbf{h}_{join}$ (Equations \ref{eq:alignScore_ours}-\ref{eq:attention}):

\vspace{-6pt}

\begin{equation}
\label{eq:alignScore_ours}
\begin{split}
e^{ij}=aligment\_model(\textbf{h}_{join}^{j},\textbf{s}^{i-1}) 
\end{split}
\end{equation}

\vspace{-18pt}

\begin{equation}
\label{eq:joint_softmax_ours}
\begin{split}
\alpha^{ij} = \dfrac{exp(e^{ij})}{\sum_{m=1}^{N+M}exp(e^{im})}
\end{split}
\end{equation}


\begin{equation}
\label{eq:attention}
\textbf{c}^{i}=\sum_{j=1}^{N+M}\alpha^{i,j}\textbf{h}_{join}^j 
\end{equation}

\noindent where $i$ is the time step on the decoder side, $j$ is the index of the hidden encoded vector. Given that the $\alpha^{i,j}$ weights form a normalized probability distribution over $j$, this model is ``forced'' to spread the weight between the \textit{src} and \textit{mt} inputs. Note that our model differs from that proposed by \cite{libovicky2017} only in that we do not employ the learnable projection matrices, $U_{c(k)}$. This is done to avoid re-weighing the contribution of the two inputs in the context vectors and, ultimately, in the predictions. More details of the proposed model and its hyper-parameters are provided in Section \ref{train_hyper}.


\section{Experiments}\label{experiments}

\subsection{Datasets}\label{dataset}

For training and evaluation we have used the WMT17 APE\footnote{http://www.statmt.org/wmt17/ape-task.html} IT domain English-German dataset. This dataset consists of 11,000 triplets for training, 1,000 for validation and 2,000 for testing. The hyper-parameters have been selected using only the validation set and used unchanged on the test set. We have also trained the model with the 12,000 sentences from the previous year (WMT16), for a total of 23,000 training triplets.

\subsection{Artificial data}\label{artificial_data}

Since the training set provided by the shared task is too small to effectively train neural networks, \cite{junczys2016} have proposed a method for creating extra, ``artificial'' training data using round-trip translations. First, a language model of the target language (German here) is learned using a monolingual dataset. Then, only the sentences from the monolingual dataset that have low perplexity are round-trip translated using two off-the-shelf translators (German-English and English-German). The low-perplexity sentences from the monolingual dataset are treated as the \textit{pe}, the German-English translations as the \textit{src}, and the English-German back-translations as the \textit{mt}. Finally, the (\textit{src}, \textit{mt}, \textit{pe}) triplets are filtered to only retain sentences with comparable TER statistics to those of the manually-annotated training data. These artificial data have proved very useful for improving the accuracy of several neural APE systems, and they have therefore been included in the WMT17 APE shared task. In this paper, we have limited ourselves to using 500K artificial triplets as done in \cite{varis2017,berard2017}. To balance artificial and manually-annotated data during training, we have resampled the official 23K triplets 10 times.

\begin{table}[t]
	\centering
	\resizebox{0.30\textwidth}{!}{\begin{tabularx}{0.7\columnwidth}{|l|X|}
			
			\hline
			\# encoders&2\\
			\hline
			encoder type&B-LSTM\\
			\hline
			encoder layers&2\\
			\hline
			encoder hidden dim&500\\
			\hline
			\# decoders&1\\
			\hline
			decoder type&LSTM\\
			\hline
			decoder layers&2\\
			\hline
			decoder hidden dim&500\\
			\hline
			word vector dim&300\\
			\hline
			attention type&\textit{general}\\
			\hline
			dropout&0.3\\
			\hline
			beam size&5\\
			\hline
			
	\end{tabularx}}
	\caption{The model and its hyper-parameters.}\label{tab:1}
	
\end{table}

\subsection{Training and hyper-parameters}\label{train_hyper}

Hereafter we provide more information about the model's implementation, its hyper-parameters, the pre-processing and the training to facilitate the reproducibility of our results. We have made our code publicly available\footnote{https://github.com/ijauregiCMCRC/Shared\_ \\ Attention\_for\_APE}.

To implement the encoder/decoder with separate encoders for the two inputs (\textit{src}, \textit{mt}) and a single attention mechanism, we have modified the open-source OpenNMT code \cite{klein2017}.  

Table \ref{tab:1} lists all hyper-parameters which have all been chosen using only training and validation data. The two encoders have been implemented using a Bidirectional Long Short-Term Memory (B-LSTM) \cite{hochreiter1997} while the decoder uses a unidirectional LSTM. Both the encoders and the decoder use two hidden layers. For the attention network, we have used the OpenNMT's \textit{general} option \cite{luong2015}.

As for the pre-processing, the datasets come already tokenized. Given that German is a morphologically rich language, we have learned the subword units using the BPE algorithm \cite{sennrich2015} only over the official training sets from the WMT16 and WMT17 IT-domain APE shared task (23,000 sentences). The number of \textit{merge} operations has been set to 30,000 under the intuition that one or two word splits per sentence could suffice. Three separate vocabularies have been used for the (\textit{src}, \textit{mt} and \textit{pe}) sentences. Each vocabulary contains a maximum of 50,000 most-frequent subword units; the remaining tokens are treated as unknown ($<$\textit{unk}$>$).

\begin{table}[t]
	\centering
	\resizebox{0.40\textwidth}{!}{\begin{tabularx}{0.88\columnwidth}{|l|l|X|}
			
			\hline
			Model&TER&BLEU\\
			\hline
			MT \cite{bojar2017}&24.48&62.49\\
			SPE \cite{bojar2017}&24.69&62.97\\
			\hline
			\cite{varis2017}&24.03&64.28\\
			\cite{berard2017}&22.81&65.91\\
			\hline
			train 11K &41.58&43.05\\
			train 23K &30.23&57.14\\
			train 23K + 500K &\textbf{22.60}&\textbf{66.21}\\
			\hline
			
	\end{tabularx}}
	\caption{Results on the WMT17 IT domain English-German APE test set.}\label{tab:2}
	
\end{table}

As mentioned in Section 4.2, we have trained our model with 500K extra triplets as in \cite{berard2017}. We have oversampled the 23K official triplets 10 times, added the extra 500K, and trained the model for 20 epochs. We have used Stochastic Gradien Descent (SGD) with a learning rate of 1 and a learning rate decay of 0.5. The learning rate decays if there are no improvements on the validation set. 

In all cases, we have selected the models and hyper-parameters that have obtained the best results on the validation set (1,000 sentences), and reported the results blindly over the test set (2,000 sentences). The performance has been evaluated in two ways: first, as common for this task, we have reported the accuracy in terms of Translation Error Rate (TER) \cite{snover2006} and BLEU score \cite{papineni2002}. Second, we present an empirical analysis of the attention weight matrices for some notable cases.

\begin{figure}[t!]
	\centering
	\includegraphics[width=0.9\linewidth]{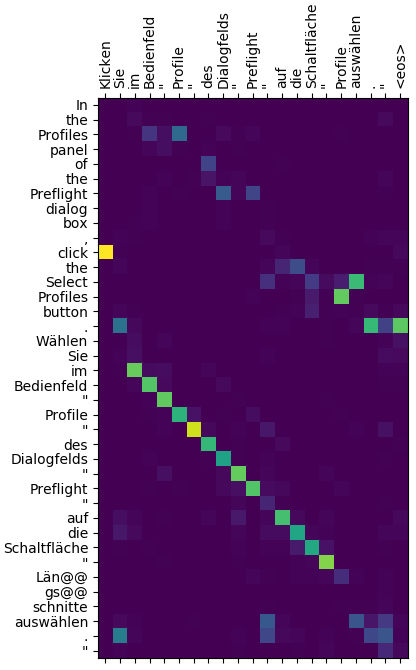}
	\caption{An example of perfect correction of an \textit{mt} sentence.}
	\label{fig:fig1}
\end{figure}

\begin{figure}[t!]
	\centering
	\includegraphics[width=0.8\linewidth]{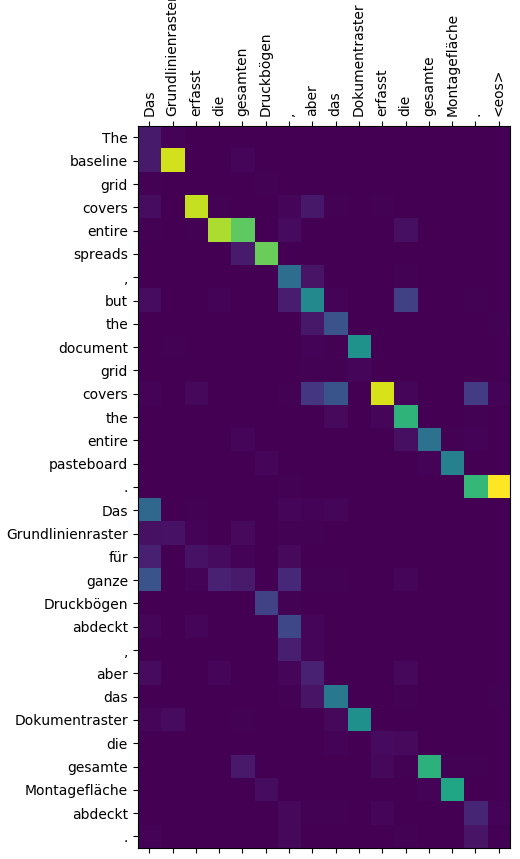}
	\caption{Partial improvement of an \textit{mt} sentence.}
	\label{fig:fig2}
\end{figure}

\begin{figure}[t!]
	\centering
	\includegraphics[width=0.9\linewidth]{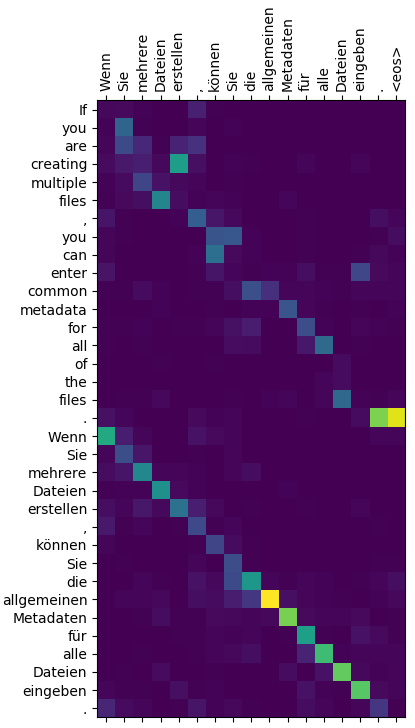}
	\caption{Passing on a correct \textit{mt} sentence.}
	\label{fig:fig3}
\end{figure}

\begin{figure}[t!]
	\centering
	\includegraphics[width=0.9\linewidth]{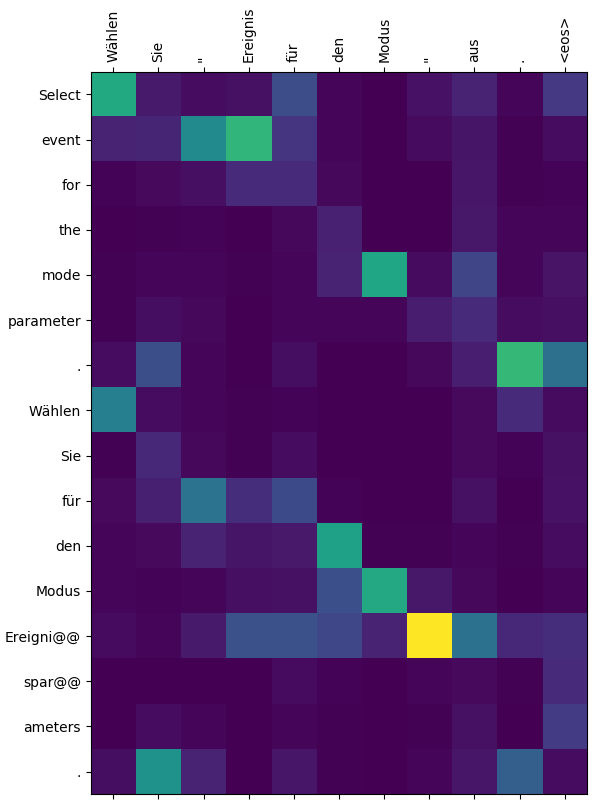}
	\caption{A completely incorrect prediction.}
	\label{fig:fig4}
\end{figure}

\subsection{Results}\label{results}

Table \ref{tab:2} compares the accuracy of our model on the test data with two baselines and two state-of-the-art comparable systems. The MT baseline simply consists of the accuracy of the \textit{mt} sentences with respect to the \textit{pe} ground truth. The other baseline is given by a statistical PE (SPE) system~\cite{simard2007} chosen by the WMT17 organizers. Table \ref{tab:2} shows that when our model is trained with only the 11K WMT17 official training sentences, it cannot even approach the baselines. Even when the 12K WMT16 sentences are added, its accuracy is still well below that of the baselines. However, when the 500K artificial data are added, it reports a major improvement and it outperforms them both significantly. In addition, we have compared our model with two recent systems that have used our same training settings (500K artificial triplets + 23K manual triplets oversampled 10 times), reporting a slightly higher accuracy than both (1.43 TER and 1.93 BLEU p.p. over \cite{varis2017} and 0.21 TER and 0.30 BLEU p.p. over \cite{berard2017}). Since their models explicitly predicts edit operations rather than post-edited sentences, we speculate that these two tasks are of comparable intrinsic complexity.

In addition to experimenting with the proposed model (Equation \ref{eq:attention}), we have also tried to add the projection matrices of the flat attention of \cite{libovicky2017} (Equation \ref{eq:flat_attention_context}). However, the model with these extra parameters showed evident over-fitting, with a lower perplexity on the training set, but unfortunately also a lower BLEU score of 53.59 on the test set. 
On the other hand, \cite{chatterjee2017} and other participants of the WMT 17 APE shared task~\footnote{http://www.statmt.org/wmt17/ape-task.html} were able to achieve higher accuracies by using 4 million artificial training triplets. Unfortunately, using such a large dataset sent the computation out of memory on a system with 32 GB of RAM. Nonetheless, our main goal is not to establish the highest possible accuracy, but rather contribute to the interpretability of APE predictions while reproducing approximately the same accuracy of current systems trained in a comparable way.

For the analysis of the interpretability of the system, we have plotted the attention weight matrices for a selection of cases from the test set. These plots aim to show how the shared attention mechanism shifts the attention weights between the tokens of the \textit{src} and \textit{mt} inputs at each decoding step. In the matrices, the rows are the concatenation of the \textit{src} and \textit{mt} sentences, while the columns are the predicted \textit{pe} sentence. To avoid cluttering, the ground-truth \textit{pe} sentences are not shown in the plots, but they are commented upon in the discussion. Figure \ref{fig:fig1} shows an example where the \textit{mt} sentence is almost correct. In this example, the attention focuses on passing on the correct part. However, the start (\textit{W\"{a}hlen}) and end (\textit{L\"{a}ngsschnitte}) of the \textit{mt} sentence are wrong: for these tokens, the model learns to place more weight on the English sentence (\textit{click} and \textit{Select Profiles}). The predicted \textit{pe} is eventually identical to the ground truth.

Conversely, Figure \ref{fig:fig2} shows an example where the \textit{mt} sentence is rather incorrect. In this case, the model learns to focus almost completely on the English sentence, and the prediction is very aligned with it. The predicted \textit{pe} is not identical to the ground truth, but it is significantly more accurate than the \textit{mt}. Figure \ref{fig:fig3} shows a case of a perfect \textit{mt} translation where the model simply learns to pass the sentence on word by word. Eventually, Figure \ref{fig:fig4} shows an example of a largely incorrect \textit{mt} where the model has not been able to properly edit the translation. In this case, the attention matrix is scattered and defocused.

In addition to the visualizations of the attention weights, we have computed an attention statistic over the test set to quantify the proportions of the two inputs. At each decoding time step, we have added up the attention weights corresponding to the \textit{src} input ($\alpha^i_{src} = \sum_{j=1}^N \alpha^{ij}$) and those corresponding to the \textit{mt} ($\alpha^i_{mt} = \sum_{j=N+1}^{N+M} \alpha^{ij}$). Note that, obviously, $\alpha^i_{src} + \alpha^i_{mt} = 1$. Then, we have set an arbitrary threshold, $t = 0.6$, and counted step $i$ to the \textit{src} input if $\alpha^i_{src} > t$. If instead $\alpha^i_{src} < 1 - t$, we counted the step to the \textit{mt} input. Eventually, if $1 - t \le \alpha^i_{src} \le t$, we counted the step to both inputs. Table \ref{tab:3} shows this statistic. Overall, we have recorded 23\% decoding steps for the \textit{src},  45\% for the \textit{mt} and 31\% for both. It is to be expected that the majority of the decoding steps would focus on the \textit{mt} input if it is of sufficient quality. However, the percentage of focus on the \textit{src} input is significant, confirming its usefulness.

\begin{table}[t]
	\centering
	\resizebox{0.3\textwidth}{!}{\begin{tabularx}{0.6\columnwidth}{|l|X|}
			
			\hline
			\textbf{Sentence}&\textbf{Focus}\\
			\hline
			\textit{src}&23\%\\
			\hline
			\textit{mt}&45\%\\
			\hline
			Both&31\%\\
			\hline
			
	\end{tabularx}}
	\caption{Percentage of the decoding steps with marked attention weight on either input (\textit{src}, \textit{mt}) or both.}\label{tab:3}
	
\end{table}

\section{Conclusion}\label{conclusion}

In this paper, we have presented a neural APE system based on two separate encoders that share a single, joined attention mechanism. The shared attention has proved a key feature for inspecting how the selection shifts on either input, (\textit{src} and \textit{mt}), at each decoding step and, in turn, understanding which inputs drive the predictions. In addition to its easy interpretability, our model has reported a competitive accuracy compared to recent, similar systems (i.e., systems trained with the official WMT16 and WMT17 data and 500K extra training triplets). As future work, we plan to continue to explore the interpretability of contemporary neural APE architectures.

\section{Acknowledgements}

We would like to acknowledge the financial support received from the Capital Markets Cooperative Research Centre (CMCRC), an applied research initiative of the Australian Government.

\balance

\bibliography{acl2018}
\bibliographystyle{acl_natbib}

\end{document}